\documentclass[10pt,twocolumn,letterpaper]{article}

\usepackage{cvpr}
\usepackage{times}
\usepackage{epsfig}
\usepackage{amsmath}
\usepackage{amssymb}
\usepackage{bm}
\usepackage{arydshln}
\usepackage{xfrac}

\usepackage[utf8]{inputenc} 
\usepackage[T1]{fontenc}    
\usepackage{url}            
\usepackage{booktabs}       
\usepackage{amsfonts}       
\usepackage{amsthm,bm}
\usepackage{nicefrac}       
\usepackage{microtype}      
\usepackage{floatrow}
\newfloatcommand{capbtabbox}{table}
\usepackage{blindtext}
\usepackage{multirow}
\usepackage{enumitem}
\usepackage[linesnumbered,ruled,lined,boxed]{algorithm2e}
\usepackage{babel}
\usepackage[x11names]{xcolor}
\usepackage{sidecap}
\usepackage{subfig}

\usepackage[breaklinks=true,bookmarks=false]{hyperref}
\cvprfinalcopy 

\def\cvprPaperID{10004\vspace{-0.4cm}} 

\ifcvprfinal\fi
            
\begin{document}

\title{Extracurricular Learning:\\ Knowledge Transfer Beyond Empirical Distribution}

\colorlet{myGreen}{gray!40!green}
\colorlet{myRed}{gray!20!red}
\newcommand{\worse}[1]{\textcolor{myRed}{(-#1)}}
\newcommand{\better}[1]{\textcolor{myGreen}{(+#1)}}
\newcommand{\rbetter}[1]{\textcolor{myGreen}{(-#1$\%$)}}
\newcommand{\rworse}[1]{\textcolor{myRed}{(+#1$\%$)}}
\newcommand{\new}[1]{\textcolor{black}{#1}}
\newcommand{\mojan}[1]{\textcolor{black}{#1}}
\newcommand{\hadi}[1]{\textcolor{black}{#1}}
\newcommand{\oncelchange}[1]{\textcolor{black}{#1}}
\newcommand{\oncelcomment}[1]{\textcolor{black}{#1}}
\newcommand{\redsout}[1]{\unskip}
\newcommand*\circled[1]{\tikz[baseline=(char.base)]{
            \node[shape=circle,fill,inner sep=2pt] (char) {\textcolor{white}{#1}};}}

\author{Hadi Pouransari\\
Apple\\
{\tt\small mpouransari@apple.com}
\and
Mojan Javaheripi\\
UCSD\thanks{Work is done while doing internship at Apple.}\\
{\tt\small mojavahe@ucsd.edu}
\and
Vinay Sharma\\
Apple\\
{\tt\small sharma.vinay@apple.com}
\and
Oncel Tuzel\\
Apple\\
{\tt\small ctuzel@apple.com}

}

\maketitle

\begin{abstract}
Knowledge distillation has been used to transfer knowledge learned by a sophisticated model (teacher) to a simpler model (student). This technique is widely used to compress model complexity.
However, in most applications the compressed student model suffers from an accuracy gap with its teacher.
We propose extracurricular learning, a novel knowledge distillation method, that bridges this gap by (1) modeling student and teacher output \new{distributions}; (2) sampling examples from \new{an approximation to the} underlying data distribution; and (3) matching student and teacher output distributions \new{over this extended set including uncertain samples}.
We conduct \new{rigorous} evaluations on regression and classification tasks and show that compared to the \hadi{standard} knowledge distillation, extracurricular learning reduces the gap by \mojan{46$\%$ to 68$\%$}. This leads to major accuracy improvements compared to the empirical risk minimization-based training for various recent neural network architectures: \mojan{16$\%$} regression error reduction on the MPIIGaze dataset, \mojan{+3.4$\%$ to +9.1$\%$} improvement in top-1 classification accuracy on the CIFAR100 dataset, and \mojan{+2.9$\%$} top-1 improvement on the ImageNet dataset.
\end{abstract}

\section{Introduction}
Training an accurate model in a supervised learning setup requires a large model capacity and a large labeled dataset.
In practice, both requirements cannot be perfectly satisfied: we have limited labeled data, and model size is bounded \hadi{by} the computational budget \hadi{that is} determined by the hardware that runs the model.
Knowledge transfer/distillation and data augmentation methods have been developed to address the challenges with computational cost and data scarcity. We briefly discuss these methods, which are also the building blocks of this work.

\noindent\textbf{Knowledge distillation.} Overparameterized neural networks learn better representations that lead to better generalization accuracy~\cite{allen2019learning}. For example, both the PyramidNet-110 model~\cite{han2017deep} and the larger PyramidNet-200 model achieve perfect accuracy on the CIFAR100~\cite{krizhevsky2009learning} training set, while the latter has 3$\%$ higher generalization accuracy. This motivated transferring the ``knowledge'' encoded in the more accurate larger model to the smaller one.
Knowledge Distillation~\cite{bucilu?2006model,hinton2015distilling} (KD) established an important mechanism through which one model (typically of higher capacity, called teacher) can train another model (typically a smaller model that satisfies the computational budget, called student). KD has been implemented in many machine learning tasks, for example image classification~\cite{hinton2015distilling}, object detection~\cite{chen2017learning,wang2019distilling}, video labeling~\cite{zhou2018edf}, natural language processing~\cite{tang2019distilling,mukherjee2019distilling,sun2019patient,liu2019improving,turc2019well}, and speech recognition~\cite{chebotar2016distilling,takashima2018investigation,liu2019end}.
%
%

The idea of KD is to encourage the student to imitate teacher's \new{behavior} over \new{a set of data points, called transfer-set}.
For example, in classification, the teacher's output includes not only the correct class index (the argmax of softmax generated probabilities), but also additional information regarding similarities to other classes.

The amount of additional information can be quantified by the entropy of the class probabilities produced by the teacher. A teacher with small training loss produces low entropy outputs over the dataset, making KD less effective.
Previously proposed remedies for this issue include matching the logits of student and teacher~\cite{bucilu?2006model},  increasing the entropy by smoothing teacher's output~\cite{hinton2015distilling}, encouraging the student to match its intermediate feature maps to that of the teacher~\cite{romero2014fitnets}, or explicitly training a teacher with high entropy outputs~\cite{pereyra2017regularizing}.
\new{In this work, we show that using a transfer-set containing uncertain examples along with modeling uncertainty of the teacher addresses this issue, and help bridging the gap between the student and teacher accuracies.}

\noindent\textbf{Data augmentation.} Lack of sufficient labeled data is another challenge in supervised learning. There are several data augmentation approaches to tackle this challenge. These methods exploit domain knowledge to transform training examples to generate more data~\cite{simard2003best,krizhevsky2012imagenet,devries2017improved}, learn a data generation policy \cite{cubuk2019randaugment,cubuk2019autoaugment,ho2019population,lim2019fast,zoph2019learning,wei2020circumventing}, augment the intermediate features of the model~\cite{gastaldi2017shake,yamada2018shakedrop}, or find difficult examples using adversarial training~\cite{xie2019adversarial}. Some of the recent methods~\cite{zhang2017mixup,verma2018manifold,yun2019cutmix,hendrycks2019augmix,gong2020maxup} mix two or more data points from the empirical distribution to generate new data points.  Alternatively, instead of manually designed transformations, generative models~\cite{oord2016pixel,oord2016,kingma2013auto,goodfellow2014generative} could be utilized to sample new training examples. Here, we \redsout{also} use samples from \new{an approximation to the} data distribution to construct an improved KD algorithm. Note that unlike classical data augmentation methods that require labels for the augmented data, in our KD framework we only need unlabeled samples.

\hadi{To bridge the gap between the teacher and the student models, in this work} we present a novel KD method,  Extracurricular Learning (XCL). Our method is motivated by the following two arguments. First, modeling the output distribution (rather than point estimates) of teacher is important for knowledge transfer as it provides additional information for student. For regression, we explicitly model the output distribution of the teacher as a Gaussian and transfer it to the student model. For classification, the output is already encoded as a categorical distribution. Second, if student exactly matches the teacher's output on the entire input domain, we are guaranteed to bridge the \hadi{accuracy} gap. This is \new{infeasible} in practice \new{due to the student's limited capacity and optimization imperfections.}
\new{As such, we propose to match the student and teacher on an extended transfer-set beyond the empirical distribution, particularly where the teacher has high uncertainty. 
We investigate various approximations of the data distribution to synthesize new examples (the extracurricular material) with high teacher uncertainty for KD.}
Our main contributions are:
\begin{itemize}[leftmargin=*,itemsep=0.1em]
\item \new{We empirically show that uncertain samples and uncertainty estimation result in significant improvements in KD generalization accuracy.}
\item We introduce XCL: a combination of modeling student and teacher output distributions, sampling (uncertain) data points from an approximate data distribution, and KD over this extended transfer-set. XCL does not require additional unlabeled samples and \hadi{or} hyper-parameter tuning.
\item XCL reduces the accuracy gap between the student and the teacher by \mojan{46$\%$ to 68$\%$} compared to \new{standard} KD. 
Compared to \new{best practice} supervised learning baselines, XCL provides \mojan{\bf 16$\%$} reduction in regression error on the MPIIGaze dataset and \mojan{\bf +3.4$\%$} (PyramidNet), \mojan{\bf +4.6$\%$} (ResNet), \mojan{\bf +9.1$\%$} (BinaryNet) top-1 classification accuracy improvement on the CIFAR100 \hadi{dataset}, and \mojan{\bf +2.9$\%$} (ResNet) on the ImageNet \hadi{dataset}.
\end{itemize}

\section{Preliminaries}
In supervised learning, we seek parameters $\bm{\theta}$ of a \new{parameterized} function $f_{\bm{\theta}}$ (e.g., weights of a neural network) to minimize the expected risk:
\begin{equation}\label{eqn:expected_risk}
\setlength{\abovedisplayskip}{5pt}
\setlength{\belowdisplayskip}{5pt}
\min_{\bm{\theta}}  \mathbb{E}_{(\bm{x},\bm{y})\sim p} [ l(f_{\bm{\theta}}(\bm{x}), \bm{y}) ],
\end{equation}
where $p(\bm{x},\bm{y})$ is the joint distribution of (example, label) pairs and $l(\cdot)$ is the loss function determining how close $f_{\bm{\theta}}(\bm{x})$ and $\bm{y}$ are. For almost every practical problem, $p$ is not available, yet a finite set of training data points $\mathcal{D} = \{\bm{x}_i, \bm{y}_i \}_{i=1}^n$ is given. The empirical risk approximation of (\ref{eqn:expected_risk}) \hadi{substitutes} $p$ with empirical distribution $p_{\delta}=1/n \sum_{i=1}^n \delta(\bm{x}=\bm{x}_i, \bm{y}=\bm{y}_i)$,
where $\delta(\bm{x}=\bm{x}_i, \bm{y}=\bm{y}_i)$ is a Dirac mass function located at $(\bm{x}_i,\bm{y}_i)$. This leads to the Empirical Risk Minimization (ERM):
\begin{equation}\label{eqn:empirical_risk}
\setlength{\abovedisplayskip}{5pt}
\setlength{\belowdisplayskip}{5pt}
\min_{\bm{\theta}} \frac{1}{n} \sum_i  l(f_{\bm{\theta}}(\bm{x}_i), \bm{y}_i) 
\end{equation}
In KD~\cite{hinton2015distilling}, a student model $f_{\theta}$ is encouraged to match the output of a teacher $\tau$ on the training set:
\begin{equation}\label{eqn:empirical_kd_risk}
\setlength{\abovedisplayskip}{5pt}
\setlength{\belowdisplayskip}{5pt}
\min_{\bm{\theta}} \frac{1}{n} \sum_i  l(f_{\bm{\theta}}(\bm{x}_i), \tau(\bm{x}_i))
\end{equation}
$\tau$ in (\ref{eqn:empirical_kd_risk}) can be a single more powerful model or an ensemble of several models. 
In the original KD~\cite{hinton2015distilling} an average of losses in (\ref{eqn:empirical_risk}) and (\ref{eqn:empirical_kd_risk}) is used.

KD is widely studied for the classification task, where $\bm{y}_i$ is a one-hot vector that indicates the true class of $\bm{x}_i$.
The teacher output $\tau(\bm{x}_i)$, however, is a soft-label.
Components of $\tau(\bm{x}_i)$ encode similarities of $\bm{x}_i$ to other classes~\cite{hinton2015distilling}, which encapsulate additional information compared to $\bm{y}_i$. Hence, training a model with soft-labels from a stronger teacher instead of one-hot labels leads to accuracy gain.

\section{Effect of Uncertainty on KD \hadi{Performance}}\label{sec:observe}
\new{In this section, we present two sets of experiments that illustrate \hadi{the} importance of (1) uncertain data points \hadi{in the transfer-set}; and (2) modeling teacher and student uncertainties on KD.}



\noindent\textbf{Uncertain samples from data distribution are important for KD.} 
\new{ 
In the first experiment, we randomly divide the data available in the training set into two disjoint subsets represented by $\mathcal{A}$ and $\mathcal{B}$. $\mathcal{A}$ corresponds to half of the data that is used to train the teacher while $\mathcal{B}$ denotes the held-out set.
We quantify data uncertainty (aleatoric uncertainty) using the entropy of predictions generated by the teacher.
We then split $\mathcal{B}$ into two equally sized distjoint subsets $\mathcal{B} = \mathcal{H} \cup \mathcal{L}$, corresponding to samples with high and low uncertainties, respectively. 
}



\new{For a soft-label $\bm{y}$ the normalized entropy is defined as}
\begin{equation}\label{eqn:entropy}
\hat{H}(\bm{y}) = -\sum_{j=1}^c \bm{y}^j \log \bm{y}^j / \log c,   
\end{equation}
\new{where superscript $j$ refers to the $j$'th component of the vector, and $c$ is the number of classes. $\hat{H}(\bm{y})$ varies between $0$ and $1$, and denotes label uncertainty for a sample. It can also be interpreted as the amount of additional information encoded in soft labels compared to one-hot labels.}

\new{
In Table~\ref{tab:aleatoric}, we report average normalized entropies of the transfer-sets, and validation accuracies of students models trained using ERM (using one-hot ground truth labels) and KD (using soft-labels from the teacher model).
For both datasets, we observe a transfer-set with higher uncertainty results in more effective KD, and therefore higher validation accuracy for student.
This trend does not hold for ERM.}
\new{Note that we can artificially create a transfer-set containing uncertain \hadi{samples}, for example, a transfer-set consisting of Gaussian noise. However, \hadi{uncertain samples that are not from the data distribution are not suitable for KD.} Please see Table~\ref{tab:qx} for this experiment.}

\new{As observed, for an effective KD we need \emph{uncertain samples from the underlying data distribution}. Our intuition is that uncertain samples are located close to the decision boundaries of the teacher model. Therefore, they \hadi{better} characterize teacher's decision boundaries compared to samples with low uncertainty (that are far away from the decision boundaries).}

\begin{table}[tb]
\centering
\resizebox{0.9\columnwidth}{!}{
\begin{tabular}{lcccc}
\hline
dataset  &
\begin{tabular}{@{}c@{}} transfer\\ set \end{tabular} & 
\begin{tabular}{@{}c@{}} entropy\\ ($\%$)  \end{tabular}& 
ERM ($\%$)   & KD ($\%$)\\ \hline \hline

\multirow{2}{*}{CIFAR100}& 
$\mathcal{H}$&
56.4&
52.7&
70.7\\

& $\mathcal{L}$&
5.5&
61.9&
65.3 \\


\hline

\multirow{2}{*}{ImageNet}& 
$\mathcal{H}$&
23.1&
65.2&
73.4 \\

& $\mathcal{L}$&
1.8&
65.2&
70.7 \\


\hline
\end{tabular}
\caption{\new{Effect of uncertainty of the transfer-set on student's top-1 validation accuracy. Teacher's top-1 accuracy is $74.7\%$ and $76.1\%$ for CIFAR100 and ImageNet, respectively.}}
\label{tab:aleatoric}}
\end{table}

\noindent\textbf{Uncertainty modeling is important for KD.} \new{In the second experiment, we analyze the effect of additional information contained in teacher's output distribution on KD using an extreme transfer-set.
Let $\mathcal{Z} \subset \mathcal{B}$ be the set of all samples for which the teacher is incorrect\footnote{$|\mathcal{Z}|$ constitutes $\sim$12$\%$ of the training data for both benchmarks.}, \hadi{i.e.,} the argmax of teacher's prediction points to a wrong class.
In Table~\ref{tab:zero}, we present results of training student models using ground truth labels (ERM) versus soft-labels from teacher (KD).
Using ERM over $\mathcal{Z}$ results in very low accuracy. In contrast, when we use soft-labels from the teacher, the student model achieves surprisingly high accuracy. This illustrates that uncertainty modeling (captured through teacher's output distribution) is crucial for KD\hadi{, particularly when} using difficult \hadi{uncertain} examples.}

\setlength{\textfloatsep}{1pt}
\begin{table}[tb]
\centering
\resizebox{0.9\columnwidth}{!}{
\begin{tabular}{lcccc}
\hline
dataset  &  entropy ($\%$)& ERM ($\%$) & KD ($\%$)\\ \hline \hline

CIFAR100& 
58.3&
14.2&
61.3\\ \hline

ImageNet&
26.7&
15.6&
58.5 \\
\hline
\end{tabular}
\caption{\new{Student top-1 validation accuracy trained over a transfer-set on which the teacher top-1 accuracy is 0$\%$.}}
\label{tab:zero}}
\end{table}

\section{Extracurricular Learning (XCL)}
Motivated by the observations discussed in Section~\ref{sec:observe}, we develop an improved KD algorithm, \hadi{dubbed} extracurricular learning (XCL), for regression and classification tasks \hadi{which utilizes} uncertainty estimation and extended transfer-sets. 

\subsection{KD Using Uncertainty Modeling}\label{sec:regress}
Uncertainty estimation is important for effective knowledge transfer since (1) it provides student with not only point estimates of teacher's output, but also the full distribution (\hadi{thus,} more is learned from the teacher); and (2) it prevents over-penalizing student on samples that teacher is not confident \hadi{about}.

\noindent\textbf{Uncertainty estimation for classification.} 
In the standard classification \hadi{task}, uncertainty is already modeled, where the teacher's output is as a categorical distribution capturing the conditional probability of the label $\bm{y}$ given $\bm{x}$.
The label distribution provides the student with the uncertainty associated to data points. Examples of data uncertainties are when different classes are present in an image, or when there is ambiguity due to occlusion. The student model is trained to minimize the average Kullback-Leibler (KL) divergence from its predicted categorical distribution to that of the teacher.

\noindent\textbf{Uncertainty estimation for regression.}
A trivial extension of KD to regression tasks is replacing the ground truth labels (regression targets) with \hadi{the teacher} predictions. Recently, Saputra \emph{et al.}~\cite{saputra2019distilling} explored some variations of KD for regression. However, these methods lack uncertainty modeling, which we show is a key property for effective KD. We introduce a KD algorithm for regression incorporating uncertainty estimation. 

We model the heteroscedastic uncertainties (uncertainties that depend on each example $x_i$) for regression tasks, similar to~\cite{nix1994estimating,kendall2017uncertainties}. Specifically, for a data point $\bm{x}_i$ we assume the model outputs $f_{\bm{\theta}}(\bm{x}_i) = (\mu_i, \sigma_i^2)$ approximating the conditional probability $p(y|\bm{x}_i)$ with a Gaussian $\mathcal{N}(\mu_i, \sigma_i^2)$. Hence, $\mu_i$'s regress $y_i$'s and $\sigma_i$'s indicate the uncertainties. \hadi{To estimate $\mu_i$'s and $\sigma_i$'s without having access to ``uncertainty labels'', we learn model parameters by minimizing the Negative Log Likelihood (NLL) loss}:
\begin{equation}\label{eqn:loglike}
\setlength{\abovedisplayskip}{5pt}
\setlength{\belowdisplayskip}{5pt}
\begin{split}
l(f_{\bm{\theta}}(\bm{x}_i), y_i) & = \frac{1}{2 \sigma_i^2} \| \mu_i - y_i \|_2^2 + \frac{1}{2} \log \sigma_i^2 \\
& = \frac{1}{2} \exp(-s_i) \| \mu_i - y_i \|_2^2 + \frac{1}{2}s_i
\end{split}
\end{equation}
In practice, the model predicts log variance $s_i = \log \sigma_i^2$ \hadi{for numerical stability}. This can be simply implemented by adding an additional output to the last layer of a neural network. The computational overhead for uncertainty estimation is negligible.

In our framework, both student and teacher predict the output conditional distribution as Gaussians: $\mathcal{N}(\mu_i, \sigma_i^2)$ and  $\mathcal{N}(\mu_i^{\tau}, {\sigma_i^{\tau}}^2)$, respectively.
 We train the teacher using (\ref{eqn:loglike}). We \hadi{then} distill the teacher's knowledge to the student by minimizing the KL divergence between two Gaussians
\begin{equation}\label{eqn:reg-kl}
\setlength{\abovedisplayskip}{2pt}
\setlength{\belowdisplayskip}{2pt}
\begin{aligned}
&l(f_{\bm{\theta}}(\bm{x}_i), \tau(\bm{x}_i)) = D_{\text{KL}} (\mathcal{N}(\mu_i^{\tau}, {\sigma_i^{\tau}}^2) ~\|~ \mathcal{N}(\mu_i, \sigma_i^2 ))\\
&=\frac{1}{2} \left[ \exp(s_i^{\tau} - s_i) + \exp(-s_i) \| \mu_i^{\tau} - \mu_i \|_2^2- (s_i^{\tau} - s_i) - 1 \right]
\end{aligned}
\end{equation}
over the transfer-set.

In Section \ref{sec:gaze}, we show KD for regression using the loss in (\ref{eqn:reg-kl}) improves student accuracy significantly compared to \hadi{methods that do not account} for uncertainties.

\subsection{Data Distribution Modeling}\label{sec:dist}
XCL extends knowledge transfer to data points beyond the empirical distribution. 
We approximate the data distribution $p(\bm{x})$ by $q(\bm{x})$, construct a transfer-set by sampling from $q$, and  perform KD over this set:
\begin{equation}\label{eqn:expected_kd}
\setlength{\abovedisplayskip}{5pt}
\setlength{\belowdisplayskip}{5pt}
\min_{\theta}  \mathbb{E}_{\bm{x}\sim q} [ l(f_{\bm{\theta}}(\bm{x}), \tau(\bm{x})) ]
\end{equation}
In classification, $l(\cdot)$ is the KL divergence between two categorical distributions, and in regression it is between two Gaussians as in (\ref{eqn:reg-kl}).

We can also interpret XCL as expected risk minimization~(\ref{eqn:expected_risk}) over an approximation of the joint (example, label) distribution $p(\bm{x},\bm{y})$. XCL approximates the expected risk more accurately compared to the ERM in~(\ref{eqn:empirical_risk}) by deploying a more accurate approximation of the joint distribution $p(\bm{x},\bm{y}) = p(\bm{x}) p(\bm{y}|\bm{x})$.
\hadi{Specifically, (\ref{eqn:expected_kd}) can be obtained from (\ref{eqn:expected_risk})}
if we approximate the data distribution $p(\bm{x})$ by $q(\bm{x})$, and the label conditional distribution $p(\bm{y}|\bm{x})$ by teacher's output distribution (Gaussian in regression and categorical in classification), and define the loss function $l(f_{\bm{\theta}}(\bm{x}), \bm{y})$ to be the negative log likelihood.

$q(x)$ can be any function that approximates the data distribution, e.g., unlabeled data, generative models~\cite{oord2016pixel,oord2016,kingma2013auto,goodfellow2014generative}, data augmentation~\cite{simard2003best,krizhevsky2012imagenet,devries2017improved},  data mixing~\cite{zhang2017mixup,verma2018manifold,yun2019cutmix,hendrycks2019augmix,gong2020maxup}, vicinal distribution~\cite{chapelle2001vicinal}, etc.

Compared to KD on the empirical distribution as in~(\ref{eqn:empirical_kd_risk}), in XCL we match student and teacher on much more data points. Specifically, when $q(\bm{x})$ is a good approximation to data distribution, we encourage the student to imitate teacher's output on high density regions, which helps transferring knowledge of the teacher to the student. 

\new{In Section~\ref{sec:observe}, we empirically showed that uncertain samples in the data distribution are more effective to distill teacher's knowledge to student. In this section, we propose two approximations to data distribution, $q(\bm{x})$ in (\ref{eqn:expected_kd}). \hadi{Our choices of $q(\bm{x})$ can be efficiently sampled from} to construct a transfer-set \hadi{that extends the empirical data distribution and includes uncertain data points}.}

\noindent\textbf{XCL-Mix: Samples from mixing in pixel space.} \new{In \hadi{our} first data distribution approximation, we model the data manifold as convex combinations of pairs of empirical data samples. To sample from this distribution, we randomly select two data points, $\bm{x}_i$ and $\bm{x}_j$ ($i,j \in \{1, n\}$), and blend them based on a random $\lambda$ as follows:}

\begin{equation} \label{eqn:mixing}
\setlength{\abovedisplayskip}{5pt}
\setlength{\belowdisplayskip}{5pt}
\begin{split}
\bm{x} =\lambda \bm{x}_i + (1-\lambda)\bm{x}_j \quad \quad &(\bm{x}_i, \bm{y}_i), (\bm{x}_j, \bm{y}_j) \sim \hat{p}_{\delta} \\
 &\lambda \sim \text{unif}[0,1]
\end{split}
\end{equation}
where $\hat{p}_{\delta}$ refers to the empirical distribution with standard augmentations and normalization. Few samples from this distribution are shown in Figure~\ref{fig:interpolation_new}. As shown, for $\lambda$ values away from 0 and 1 we sample highly uncertain points. 
This approach is similar to the augmentation method in~\cite{zhang2017mixup}, however, \hadi{here} we only sample data points, and do not use the \hadi{interpolated} labels.

\noindent\textbf{XCL-GAN: Samples from data manifold using GAN.} \new{
The above method makes a crude approximation to the data distribution by mixing samples in the pixel space. We can provide a better approximation by explicitly modeling the data manifold using a generative model and sampling from it.
For this purpose, we utilize a conditional generative adversarial network (GAN)~\cite{goodfellow2014generative} to model the data distribution and sample from it. 
The generative model $G$, given a $d-$dimensional latent variable $\bm{z}$, and a one-hot class vector $\bm{e}_i$ corresponding to class $i$, generates a sample $x=G(\bm{z};\bm{e}_i)$ from class $i$. 
\hadi{The} generative model also \hadi{provides} an explicit way to sample uncertain points when \hadi{given} a mixed class vector:}

\begin{equation} \label{eqn:ganmix}
\setlength{\abovedisplayskip}{5pt}
\setlength{\belowdisplayskip}{5pt}
\begin{split}
i, j &\sim \text{unif}\{1,c\} \\
\bm{x} =G(\bm{z}; \lambda \bm{e}_i + (1-\lambda)\bm{e}_j) \quad\quad  \lambda &\sim \text{unif}[0,1]\\
 \bm{z} &\sim \mathcal{N}_d({\bm{0}},\mathbb{I})
\end{split}
\end{equation}

\new{
In our experiments, we used BigGAN~\cite{brock2018large} trained on CIFAR100 and ImageNet datasets.
Few samples from this generative model are shown in Figure~\ref{fig:gan_interpolation}. As shown, by mixing class vectors we sample highly uncertain points. Note that, during distillation we use a union of empirical distribution and samples from the generative model as our transfer-set.}

\new{We discuss alternative sampling choices in Section~\ref{sec:altq}.}

\section{Experiments}\label{sec:exp}
We conducted experiments on regression (2D gaze estimation) and classification (CIFAR100 and ImageNet) tasks and compared XCL with ERM and alternative KD formulations. For ERM we also report results using mixing based data augmentation methods MixUp~\cite{zhang2017mixup} and CutMix~\cite{yun2019cutmix}. In MixUp, the dataset is augmented by random convex combination of data points in pixel space. For a sample $\bm{x} =\lambda \bm{x}_i + (1-\lambda)\bm{x}_j$ MixUp uses linear interpolation $\bm{y} =\lambda \bm{y}_i + (1-\lambda)\bm{y}_j$ to assign a label to $\bm{x}$, where $\bm{y}_i$ and $\bm{y}_j$ are one-hot labels corresponding to $\bm{x}_i$ and $\bm{x}_j$, respectively. In CutMix, a pair of images are blended by replacing a rectangular block of $\bm{x}_i$ with that of $\bm{x}_j$. CutMix also uses linear interpolation to assign a label to a blended image. For XCL, we separately report results using the two data distribution approximations introduced in Section~\ref{sec:dist}, denoted by XCL-Mix corresponding to sampling from blended images in pixel space, and XCL-GAN corresponding to sampling from a conditional GAN. When \hadi{comparing} XCL to KD, we also report the gap between the teacher and student accuracies (and $\%$ of gap reduction compared to standard KD). Note that, implementation details (e.g., learning rate schedule, batch size, etc.) of all of our experiments are provided in the supplementary material.


\subsection{2D Gaze Estimation}\label{sec:gaze}
We evaluate XCL on a regression task, human eye-gaze estimation, that is to predict the 2D gaze orientation vector given the image of an eye. We used the MPIIGaze dataset~\cite{zhang15_cvpr,zhang2017mpiigaze} that contains 45,000 annotated eye images of 15 persons. We followed the leave-one-person-out setup similar to the original works~\cite{zhang15_cvpr,zhang2017mpiigaze} by splitting the data to 20$\%$ validation and $80\%$ training sets. We used LeNet~\cite{lecun1998gradient} as the student and PreAct-ResNet~\cite{he2016deep} as the teacher.
Our training setup matches the accuracies reported in the original works~\cite{zhang15_cvpr,zhang2017mpiigaze}. \new{We run each experiment three times with a different \hadi{random} seed.}

\new{In Table \ref{tab:gaze}, we report the estimated angle error (in degrees). The baseline method (ERM) has 7.41 error, \hadi{which can be reduced to 6.88 using MixUP data augmentation.}
Standard KD obtains 6.65 error, that is more than $10\%$ reduction compared to ERM. We also report results using Attentive Imitation Loss (AIL)~\cite{saputra2019distilling}, a recent method that controls the extent of knowledge transfer at each data point based on teacher's error. In our experiments AIL did not improve the student accuracy compared to KD. When we incorporate \hadi{our proposed uncertainty modeling in Section~\ref{sec:regress}}
 to KD (KD+Uncertainty), the student error reduces to $6.36$, which corresponds to $36\%$ gap reduction compared to KD. The results are significantly improved using XCL (KD with uncertainty estimation and data distribution approximation), where XCL-Mix and XCL-GAN achieve {\bf 42$\%$} and {\bf 53$\%$}  teacher-student accuracy gap reductions, respectively. For all \hadi{cases in Table \ref{tab:gaze}}, we used the same teacher (with uncertainty estimation) that has an average angle error of 5.84 degrees. Note that, all methods shown in Table \ref{tab:gaze} are re-implemented, trained, and tested with identical setups.}


\begin{table}[tb]
\centering
\resizebox{0.9\columnwidth}{!}{
\begin{tabular}{llc}
  \hline
method&
angle error&
gap\\
  \hline
  \hline
ERM&
7.41$\pm$0.03&
\multirow{2}{*}{N/A}\\

ERM+MixUp~\cite{zhang2017mixup}&
6.88$\pm$0.09&
\\


\hline

KD~\cite{hinton2015distilling}&
6.65$\pm$0.03&
0.81 (-)\\

KD+AIL~\cite{saputra2019distilling}&
6.74$\pm$0.06&
{0.90 \rworse{11}} \\


KD+Uncer.&
6.36$\pm$0.05&
0.52 \rbetter{36} \\


XCL-Mix&
6.31$\pm$0.03&
0.47 \rbetter{42} \\

XCL-GAN&
{\bf 6.22$\pm$0.01}&
{\bf 0.38 \rbetter{53}} \\
  \hline
  \end{tabular}}
  \caption{Gaze angle estimation using a LeNet student model. Teacher is a PreAct-ResNet with 5.84 angle error.}
  \label{tab:gaze}
\end{table}

\begin{figure}
    \centering
    \resizebox{0.9\columnwidth}{!}{
    \includegraphics{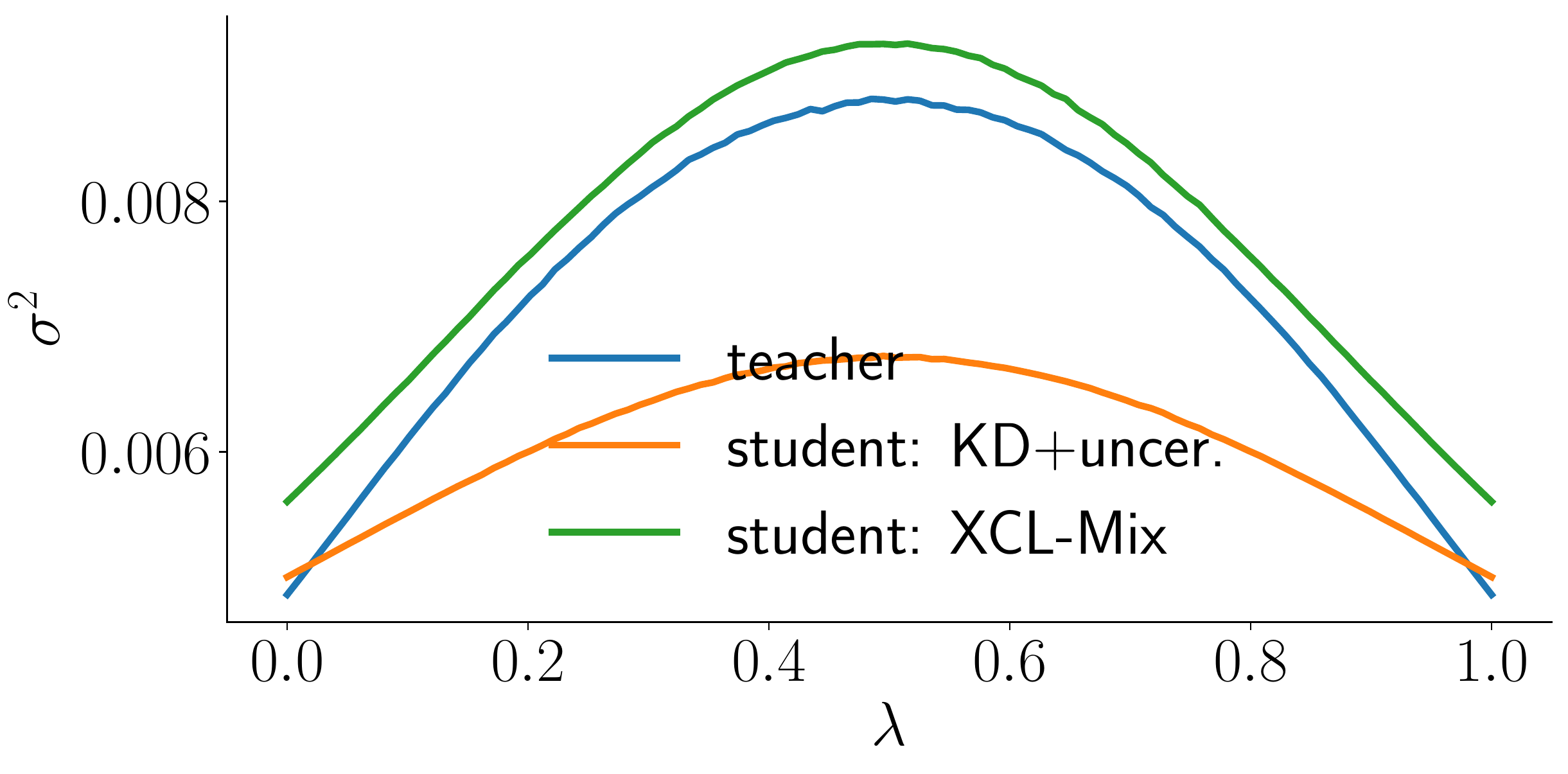}
    \caption{\new{Uncertainty estimation values for the gaze estimation task versus the mixing coefficient.}}
    \label{fig:uncertainty_regression}}
\end{figure}

In Figure \ref{fig:uncertainty_regression}, we plot the average predicted uncertainties by the teacher and two student models (trained using KD+Uncertainty and XCL-Mix), as a function of mixing coefficient $\lambda$ defined in (\ref{eqn:mixing}). As expected, as we mix the images more ($\lambda$ close to 0.5) all models predict higher uncertainties. Note that, this intuitive prediction is obtained without an explicit supervision for uncertainties. In addition, when we use XCL, the student model imitates teacher on a better data distribution approximation, and therefore has closer uncertainty estimation to the teacher on average.


\subsection{CIFAR100 Classification}\label{sec:cifar100}
We evaluate the performance of XCL for image classification task on the CIFAR100 dataset~\cite{krizhevsky2009learning}, containing 100 classes with 50k and 10k images in the training and test sets, respectively. For fair comparison, we reimplemented all \hadi{benchmarked} methods and trained with identical setups.
To compute accuracies, we first compute the median over the last 10 epochs, and then average the results over 8 independent runs with different random initializations. The standard-deviation of accuracy \hadi{for} different initializations is denoted by $\pm \mbox{std}$. 

\begin{figure*}[tb]
\centering
\includegraphics[scale=.21]{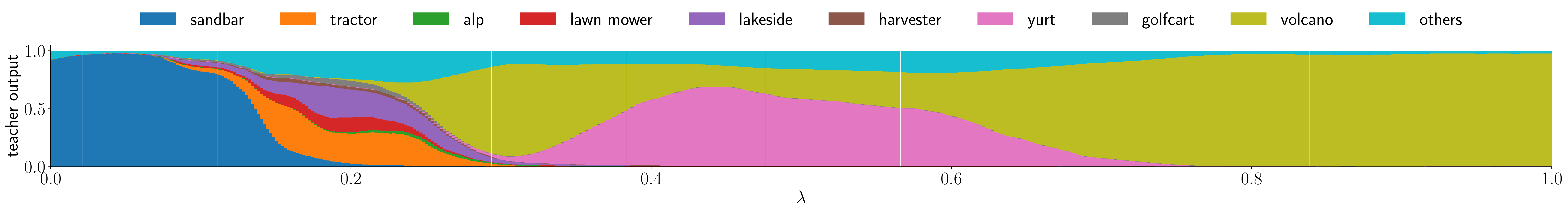}
\includegraphics[scale=.11]{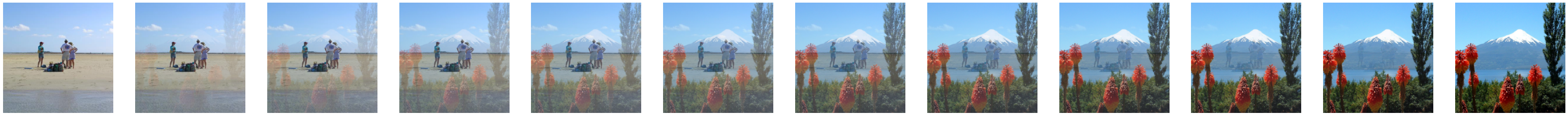}
\caption{\new{Examples of teacher outputs over a trajectory connecting two images in pixel space.}}\label{fig:interpolation_new}
\end{figure*}

{\bf ResNet-18:}  We followed the same setup as~\cite{devries2017improved} to train the ResNet-18 model~\cite{he2016deep}. We trained all methods for $2\times$ longer iterations compared to~\cite{devries2017improved}, which led to a slightly improved baseline. The reason for longer training is, by using \hadi{our} data distribution approximation we can sample infinite number of examples, therefore the training saturates later. 
To obtain an accurate teacher $\tau$, we use the ensemble method~\cite{dietterich2000ensemble} and train a committee consisting of 8 models using CutMix data augmentation~\cite{yun2019cutmix}.
The teacher's output is the ensemble average of the committee members' outputs. The ensemble model has top-1 test accuracy of $\bm{84.6\%}$. 
We explore other choices of the teacher in the supplementary material.

The results in Table \ref{tab:results_cifar100} demonstrate that both XCL-Mix and XCL-GAN significantly reduce the teacher-student accuracy gap compared to the standard KD with the same teacher (by {\bf 67 $\%$} and {\bf 46 $\%$}, respectively). This leads to major improvements over the ERM training ({\bf +4.6}$\%$) and the data mixing methods (MixUp and CutMix) that use linear interpolation for labels ({\bf $\sim$ +4}$\%$).

In Table \ref{tab:results_cifar100}, we also report the average normalized entropy, $\hat{H}$ defined in (\ref{eqn:entropy}), over the training/transfer-set of each method. 
For the standard KD method, soft-labels are obtained from a teacher trained over the empirical distribution. \hadi{Since} the teacher overfits the empirical distribution (becomes overconfident), the uncertainty of the teacher on the empirical distribution is an underestimation\footnote{As shown in the Table \ref{tab:results_cifar100}, the average entropy of the teacher over the empirical distribution is 10.5$\%$. To analyze the overfitting, we computed the same measure over the test set, which is 27.8$\%$.}.
\hadi{As a result}, using empirical distribution to distill the teacher's knowledge to the student is ineffective when the teacher overfits, which is also observed in~\cite{hinton2015distilling}.
XCL remedies the overfitting (uncertainty underestimation) problem by performing KD over an extended dataset sampled from a data distribution approximation on which teacher's uncertainties are better quantified. 



An alternative is to artificially increase $\hat{H}(y)$ by applying Label Smoothing (LS) \cite{szegedy2016rethinking}.
We applied LS to match the average normalized entropy to that of XCL's transfer-set. In Table \ref{tab:results_cifar100}, we see LS slightly improves the baseline accuracy. However, LS is worse than XCL by more than $4\%$ while having the same average entropy. We also applied smoothing by using a temperature parameter in KD~\cite{hinton2015distilling}. KD with temperature and LS required exhaustive hyper-parameter tuning. We found that using temperature can improve performance of KD by $1\%$, which is still $2\%$ worse than XCL without any parameter tuning. See the supplementary material for additional results.

\begin{table}[tb]
\centering
\resizebox{\textwidth}{!}{
  \begin{tabular}{lcccc}
  \hline
method&
\begin{tabular}{@{}c@{}} entropy\\ ($\%$) \end{tabular}&
\begin{tabular}{@{}c@{}} top-1\\ ($\%$) \end{tabular}&
\begin{tabular}{@{}c@{}} top-1\\ gap ($\%$) \end{tabular}&
\begin{tabular}{@{}c@{}} top-5\\ ($\%$) \end{tabular}\\
  \hline
  \hline
ERM&
0&
$78.5\pm$0.3&
\multirow{4}{*}{N/A}&
93.9$\pm$0.2\\

+MixUp~\cite{zhang2017mixup}&
10.8&
79.2$\pm$0.2&
&
93.9$\pm$0.2\\

+CutMix~\cite{yun2019cutmix}&
10.8&
79.3$\pm$0.2&
&
94.7$\pm$0.2\\

+LS~\cite{szegedy2016rethinking}&
28.2&
78.8$\pm$0.2&
&
93.9$\pm$0.2\\

\hline

KD~\cite{hinton2015distilling}&
10.5&
80.0$\pm$0.2&
4.6 (-)&
95.5$\pm$0.1\\

XCL-Mix&
28.2&
{\bf 83.1$\pm$0.2}&
{\bf 1.5 \rbetter{67}}&
{\bf 96.7$\pm$0.1}\\

XCL-GAN&
29.6&
82.1$\pm$0.2&
2.5 \rbetter{46}&
96.2$\pm$0.1\\

 \hline
 \end{tabular}
   \caption{Evaluation on the CIFAR100 dataset using ResNet-18. Teacher is an ensemble of 8 ResNet-18 models with 84.6$\%$ top-1 accuracy. All results are reproduced.}
  \label{tab:results_cifar100}
}
\end{table}

\begin{figure*}[t]
    \centering
    \includegraphics[width=0.75\textwidth]{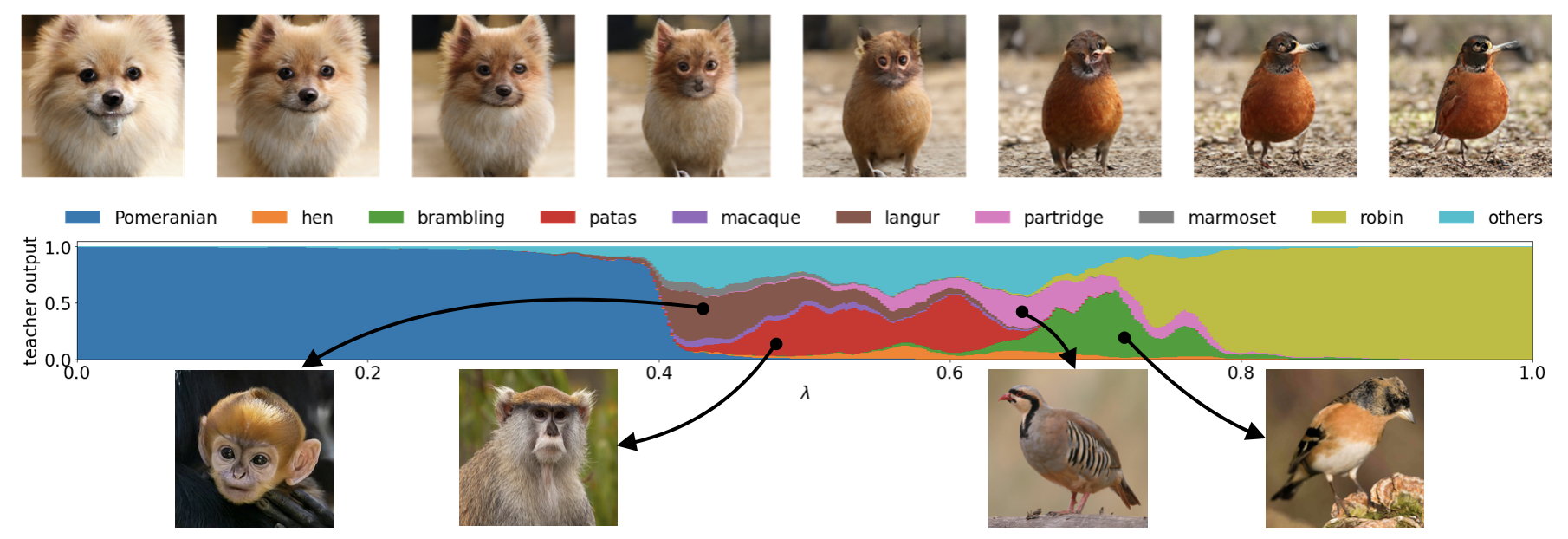}
    \caption{\new{Examples of teacher outputs over a trajectory on BigGAN manifold connecting two pure label (one-hot class vector) images. For the prominent classes we show a sample image for comparison.}}
    \label{fig:gan_interpolation}
\end{figure*}

{\bf PyramidNet-200:} We evaluate the performance of XCL on a higher capacity architecture, PyramidNet-200 \cite{han2017deep}, which obtains the state-of-the-art results on CIFAR100 dataset. We used the training setup in~\cite{yun2019cutmix}, and obtained close accuracies. The teacher is an ensemble of 8 models trained with CutMix, having a top-1 test accuracy of {\bf 87.5}$\%$. As shown in Table \ref{tab:results_cifar100_pyramid}, compared to standard KD with the same teacher, XCL significantly (by {\bf 68$\%$}) reduces teacher-student accuracy gap.

\setlength{\textfloatsep}{0.1cm}
\begin{table}[tb]
\RawFloats
\centering
\resizebox{0.7\columnwidth}{!}{
\begin{tabular}{lccc}
  \hline
method&
\begin{tabular}{@{}c@{}} top-1\\ ($\%$) \end{tabular}&
\begin{tabular}{@{}c@{}} top-1\\ gap ($\%$) \end{tabular}&
\begin{tabular}{@{}c@{}} top-5\\ ($\%$) \end{tabular}\\
  \hline
  \hline
ERM&
82.9$\pm 0.4$&
\multirow{3}{*}{N/A}&
96.3\\

+MixUp&
83.4$\pm 0.1$&
&
95.7\\

+CutMix&
84.3$\pm 0.3$&
&
96.7\\

%

\hline

KD&
83.8$\pm 0.2$&
3.7 (-)&
96.1\\

XCL-Mix&
{\bf 86.3$\pm$0.1}&
{\bf 1.2 \rbetter{68}}&
{\bf 97.6}\\
  \hline
  \end{tabular}}
    \caption{Evaluation on CIFAR100 dataset using PyramidNet-200 model. Teacher top-1 test accuracy is 87.5$\%$. 
    }
  \label{tab:results_cifar100_pyramid}
\end{table}

{\bf Quantized Networks:} We evaluate the performance of XCL to train an extremely compressed student, a Binary-Weight~\cite{courbariaux2015binaryconnect,rastegari2016xnor} ResNet-18.
This network has $~\sim 20\times$ smaller size compared to the full-precision (32-bit) model. We use the training setup as described in~\cite{martinez2020training}. Teacher is an ensemble of 8 full-precision ResNet-18 models trained with CutMix, \hadi{with} a top-1 accuracy of $84.6\%$. As shown in Table~\ref{tab:results_cifar100_binary}, compared to standard KD with the same teacher, XCL significantly (by {\bf 52$\%$}) reduces teacher-student accuracy gap.

\begin{table}[tb]
\centering
\resizebox{0.7\columnwidth}{!}{
\begin{tabular}{lccc}
  \hline
method&
\begin{tabular}{@{}c@{}} top-1\\ ($\%$) \end{tabular}&
\begin{tabular}{@{}c@{}} top-1\\ gap ($\%$) \end{tabular}&
\begin{tabular}{@{}c@{}} top-5\\ ($\%$) \end{tabular}\\
  \hline
  \hline
ERM&
70.2$\pm 0.2$&
\multirow{3}{*}{N/A}&
90.5\\

+MixUp&
72.7$\pm 0.2$&
&
90.2\\

+CutMix&
75.2$\pm 0.2$&
&
92.7\\

\hline

KD&
74.0$\pm 0.2$&
10.6 (-)&
91.9\\

XCL-Mix&
{\bf 79.5$\pm$0.1}&
{\bf 5.1 \rbetter{52}}&
{\bf 95.2}\\
  \hline
  \end{tabular}}
\caption{Evaluation on the CIFAR100 dataset using Binary-Weight ResNet-18. Teacher top-1 test accuracy is 84.6$\%$. 
}
  \label{tab:results_cifar100_binary}
\end{table}

\subsection{ImageNet Classification}
The ImageNet 2012 dataset~\cite{ILSVRC15} consists of $\sim 1.3$ million training examples and a validation set with 50,000 images from 1,000 classes. We followed the training setup in~\cite{he2019bag} and used 300 epochs for all ImageNet experiments~\cite{yun2019cutmix}. The model is a regular ResNet-101 architecture~\cite{he2016deep}. We use an ensemble of 4 ResNet-152-D~\cite{he2019bag} models trained with CutMix, having a top-1 validation accuracy of 83.3$\%$. In addition to the regular validation set of the ImageNet dataset, we evaluated the performance of the models on three recently introduced test sets for ImageNet, called ImageNetV2~\cite{recht2019imagenet} that are collected with different sampling strategies: Threshold-0.7 (V2-A), Matched-Frequency (V2-B), and Top-Images (V2-C). Results are shown in Table~\ref{tab:results_res101}.

Compared to standard KD with the same teacher, XCL-Mix reduces student-teacher validation accuracy gap by {\bf 46$\%$}. Similarly, on all other test sets, XCL obtains significant improvements compared to ERM, data mixing methods, and standard KD. 
We also report results for ResNet-50 training in the supplementary material, which shows the same trend.

\begin{table}[tb]
\centering
\resizebox{\textwidth}{!}{
\begin{tabular}{lcccccc}
  \hline
method&
\begin{tabular}{@{}c@{}}val\\ top-1\end{tabular}&
\begin{tabular}{@{}c@{}}val\\ top-1 gap\end{tabular}&
\begin{tabular}{@{}c@{}}val\\ top-5\end{tabular}&
\begin{tabular}{@{}c@{}}V2-A\\ top-1\end{tabular}&
\begin{tabular}{@{}c@{}}V2-B\\ top-1\end{tabular}&
\begin{tabular}{@{}c@{}}V2-C\\ top-1\end{tabular}\\
  \hline
  \hline
ERM&
79.0&
\multirow{3}{*}{N/A}&
94.5&
76.0&
67.5&
80.6\\
+MixUp~\cite{zhang2017mixup}&
79.7 &
&
94.8 &
77.1 &
68.2 &
81.5 \\
+CutMix~\cite{yun2019cutmix}&
80.6 &
&
95.2 &
77.1 &
69.2 &
81.7 \\
  \hline
KD~\cite{hinton2015distilling}&
80.7 &
2.6 (-)&
94.3 &
77.4 &
68.6 &
82.1\\
XCL-Mix&
{\bf 81.9 }&
{\bf 1.4 \rbetter{46} }&
{\bf 95.8 }&
{\bf 79.0 }&
{\bf 70.6 }&
{\bf 83.3 }\\
XCL-GAN&
81.6&
1.7 \rbetter{35}&
95.6&
78.4&
70.5&
82.8\\
  \hline
  \end{tabular}
  \caption{ResNet-101 evaluation ($\%$) on the ImageNet dataset. Teacher is an ensemble of 4 ResNet-152-D, with top-1 accuracy of 83.3$\%$. The std of XCL val top-1 is $\simeq 0.1$.
  }
  \label{tab:results_res101}
}
\end{table}
\section{Analysis of XCL}
In this section, we analyze alternative distribution approximations, effect of transfer-set size on distillation, and present teacher model as a non-linear interpolation method. For all experiments we use the CIFAR100 dataset and ResNet18 architecture as described in Section~\ref{sec:cifar100}.

\subsection{Alternative distribution approximations.}\label{sec:altq}
\new{The analyzed choices of data distribution approximations, $q$, are: Standard Gaussian image where each pixel is sampled from $\mathcal{N}(0,1)$; Pixel-wise Gaussian noise $\mathcal{N}(0,0.02)$ added to the empirical distribution; XCL-Mix; XCL-GAN with and without mixing class conditioning vectors (mixed-class vs. one-hot class vectors are input to GAN to generate data). As seen in Table~\ref{tab:qx}, better approximations of the data distribution, e.g., mixing methods, result in better knowledge transfer and higher student accuracy compared to uninformative distributions such as the pixel-wise Gaussian. In these experiments GAN's approximation of the data distribution is slightly worse than mixing because some modes of the distribution are not recovered by GAN and there are few visual artifacts~\cite{brock2018large}. Besides, we observe XCL-GAN performs better when mixed class vectors are used. This is consistent with the usefulness of uncertainty in the transfer-set as discussed in Section~\ref{sec:observe} for real data.}

\begin{table}[tb]
\RawFloats
\centering
\resizebox{0.9\columnwidth}{!}{
\begin{tabular}{lcc}
  \hline
sampling method&
entropy ($\%$)&
top-1 ($\%$)\\
  \hline
  \hline



Standard Gaussian Image&
53.5&
~~1.0 $\pm$ 0.0\\

Gaussian Noise Augmentation&
11.5&
79.9 $\pm$ 0.2\\


XCL-Mix&
28.2&
83.1 $\pm$ 0.2\\

XCL-GAN (no mixing)&
23.0&
81.7 $\pm$ 0.1\\

XCL-GAN&
29.6&
82.1 $\pm$ 0.2\\


  \hline
  \end{tabular}}
    \caption{Analysis of different sampling methods for XCL using the CIFAR100 dataset. Teacher has a top-1 test accuracy of 84.6$\%$. \new{For comparison, the student trained with ERM on real data has a top-1 accuracy of $78.5 \pm 0.3$.}}
  \label{tab:qx}
\end{table}

\subsection{Effect of dataset size} \label{sec:lowdata}
\begin{figure}[tb]
    \centering
    \includegraphics[width=0.62\columnwidth]{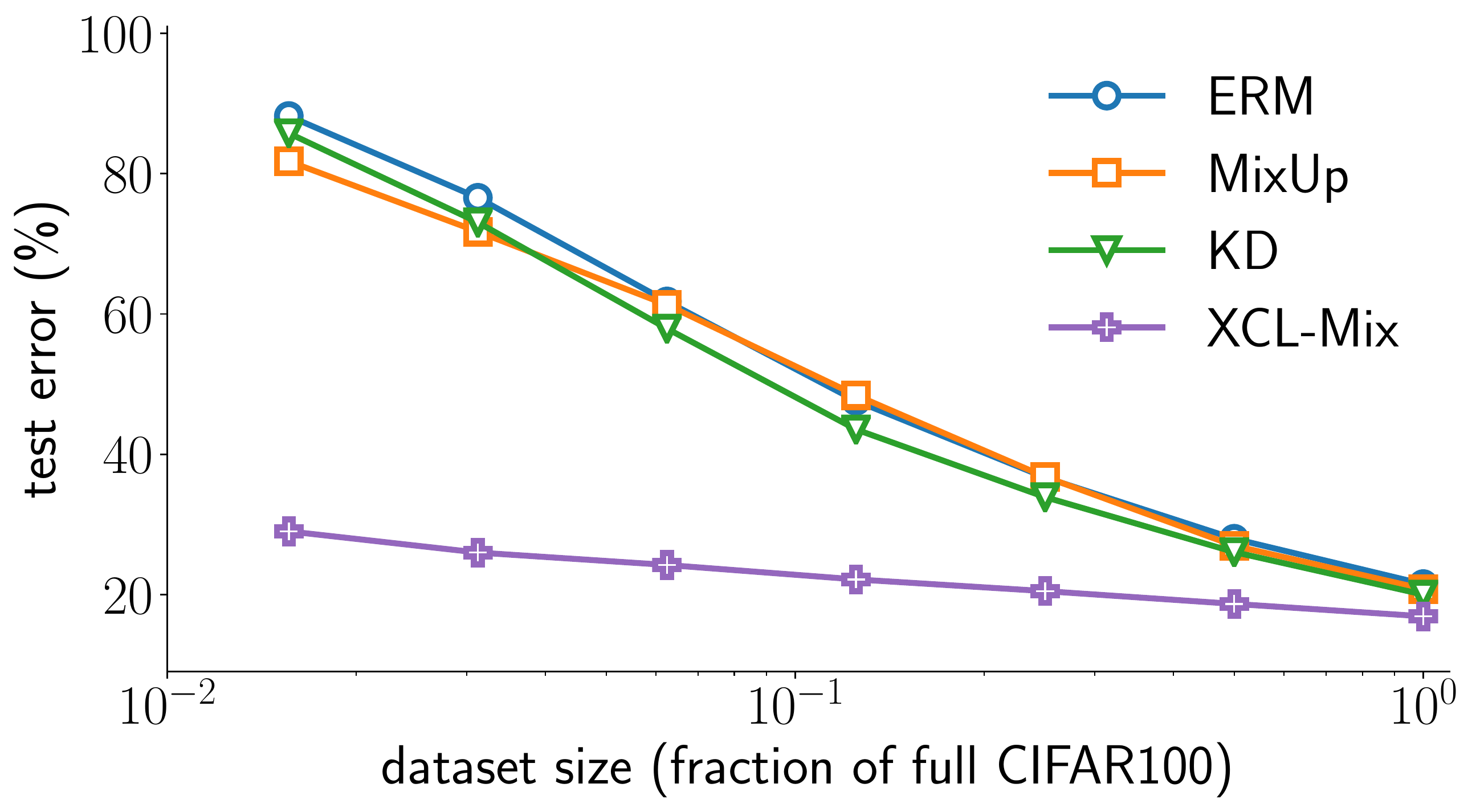}
    \caption{\new{Test error as a function of dataset size. Number of examples from all 100 classes is the same in each case.}}
    \label{fig:dataset_size}
\end{figure}

\noindent\textbf{Class balanced case.} 
\new{In Figure~\ref{fig:dataset_size} we plot test error as a function of transfer-set size when the number of samples from different classes are equal. 
We observe that XCL improvement is even more pronounced as the transfer-set size gets smaller. For example, when we use $\sfrac{1}{64}$ of the samples in the CIFAR100 as the transfer-set, ERM, MixUp, and KD obtain $11.8\%$, $14.2\%$, $18.2\%$ test accuracies, respectively. For the same setup, XCL reaches $71\%$ test accuracy, i.e., $5\times$ improvement.}

\begin{figure}
    \centering
    \resizebox{0.62\columnwidth}{!}{
    \includegraphics{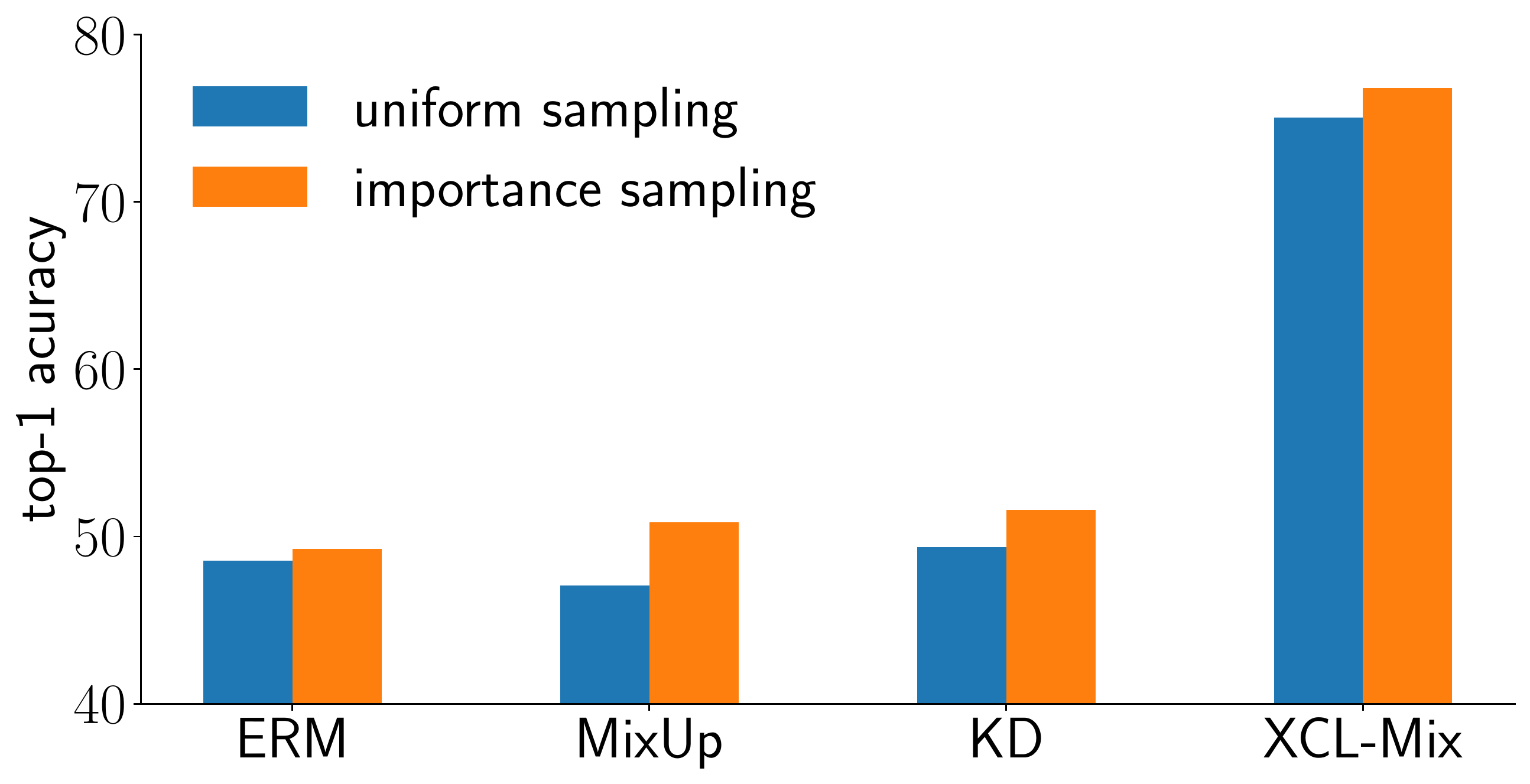}
    \caption{\new{Test accuracy for an imbalanced CIFAR100 dataset training. \hadi{The number of data samples in} 80 random classes is reduced by a factor of 10.}}
    \label{fig:imbalance}}
\end{figure}

\noindent\textbf{Class imbalanced case.} 
\new{We artificially create a class imbalanced dataset: for 80 randomly selected classes out of 100 classes of the CIFAR100 dataset, we use only $10\%$ of samples. In Figure~\ref{fig:imbalance} we report test accuracies of models trained with ERM, MixUp, KD, and XCL. XCL compensates for the class imbalance by transferring teacher's knowledge over additional examples sampled from the approximation of data distribution and outperforms all other training methods by at least a $25\%$ margin. For each training method, we also report results when training is performed using importance sampling: the sampling probability of each instance is inversely proportional to its class population. We observe improvement by up to $4\%$ when using importance sampling.}

\subsection{Teacher as a Nonlinear Interpolation Function}
\new{Several augmentation schemes such as MixUp~\cite{zhang2017mixup} and CutMix~\cite{yun2019cutmix} mix images and use linear interpolation to obtain labels of the mixed images. In this section, we argue the teacher model is a better interpolation function.}


\new{Figure~\ref{fig:interpolation_new} demonstrates the transition of teacher's output probability distribution as the mixing coefficient $\lambda$ changes.} 
Linear interpolation overconfidently assigns a label to $\bm{x}$ with similarities to $\bm{x}_i$ and $\bm{x}_j$ that add up to 1. In contrast, labels obtained by XCL for an $\bm{x}$ between $\bm{x}_i$ and $\bm{x}_j$ include similarities to all classes (not just to classes of $\bm{x}_i$ and $\bm{x}_j$). \new{For example, in Figure~\ref{fig:interpolation_new}, at $\lambda = 0.5$ the teacher predicts $\bm{x}$ to be classified as a `yurt' which is close to the visual features in the mixed image but is not equal to any of the original labels, i.e., `sandbar' or `volcano'. Similarly, in Figure~\ref{fig:gan_interpolation} we show transition between two images sampled on the image manifold by interpolating the class conditional probabilities of BigGAN (between two one-hot class vectors). As shown, the teacher output for the samples in the middle of the manifold could be different from the end-point classes (`pomeranian' and `robin'). However, qualitatively, the generated images are similar to the predicted class distribution.
} 

\vspace{-0.2cm}
\section{Other Related Works}

\new{
There are extensions of KD that match intermediate feature maps in addition to teacher outputs, e.g., FitNets~\cite{romero2014fitnets}. In our experiments, intermediate supervision produced marginal improvement compared to the standard KD ($79.3\%$), which is significantly lower than XCL (83.1$\%$).} In several works, multi-stage KD was proposed to improve both teacher and student by training a sequence of models~\cite{furlanello2018born,mirzadeh2019improved,bagherinezhad2018label}. 
Both intermediate supervision and multi-stage KD techniques are complementary to our framework, and could be incorporated to further reduce the accuracy gap.

There are several recent semi supervised learning methods~\cite{radosavovic2018data,xie2019self,berthelot2019mixmatch,sohn2020fixmatch,berthelot2019remixmatch,arazo2019pseudo,cascante2020curriculum,wang2020neural} that produce pseudo labels for unlabeled data using a model trained on a limited labeled set. The extended dataset is then used to train the target model. However, unlike XCL, these methods require additional real samples. 

\new{KD has been used to provide a fast approximation to Bayesian Neural Networks (BNNs)~\cite{balan2015bayesian,malinin2019ensemble,cui2019accelerating}. BNNs implicitly estimate the uncertainty via Monte-Carlo sampling of the network parameters. In our framework, we explicitly model and learn the output distribution of the teacher and utilize it over an extended transfer-set to reduce the KD gap.}



\section{Conclusion}
We \hadi{introduce} XCL, a framework for KD that incorporates a combination of (1) uncertainty estimation, (2) data distribution approximation, and (3) imitating the teacher output distribution using an extended transfer-set including highly uncertain points from the approximate data distribution. This results in an easy-to-use algorithm that provides the state-of-the-art accuracies for knowledge distillation (both for classification and regression \hadi{tasks}) without need for additional dataset or hyper-parameter tuning. Experiments on MPIIGaze, CIFAR100, and ImageNet datasets \hadi{show} that XCL achieves state-of-the-art accuracies for KD.



{\small
\bibliographystyle{ieee_fullname}
\bibliography{main}
}

\clearpage

\appendix
\section{Training details}
\subsection{Gaze Estimation}\label{sec:implementation:gaze}
 We used the MPIIGaze dataset~\cite{zhang15_cvpr,zhang2017mpiigaze} that contains 45,000 annotated eye images of 15 persons (3,000 images per person divided equally between left and right eyes). We followed the leave-one-person-out evaluation process similar to the original works~\cite{zhang15_cvpr,zhang2017mpiigaze}. We split the data to 20$\%$ validation and $80\%$ training sets (3 randomly selected persons are held-out for validation and 12 persons for training). We run each experiment three times with a different random seed and report average error.  We followed the implementation of original works~\cite{zhang15_cvpr,zhang2017mpiigaze} for training
, and used two existing architectures for student and teacher: the student is a 4-layer LeNet~\cite{lecun1998gradient} and the teacher is a 9-layer PreAct-ResNet~\cite{he2016deep} trained with MixUp. The models output a two-dimensional vector that predicts the gaze vector. When estimating the uncertainty, we use an isotropic Gaussian $\mathcal{N}(\mu, \sigma \mathbb{I})$ to model the output distribution. Therefore, the network output is three-dimensional. We set weight decay to $10^{-4}$, learning rate to $10^{-4}$ for LeNet and $10^{-3}$ for ResNet that is decayed by a factor of 10 after 30 and 36 epochs. All models are trained using ADAM optimizer~\cite{kingma2014adam} for 40 epochs with 0.9 momentum and batch-size of 32.

\subsection{ResNet-18 on CIFAR100}
We followed the same setup as~\cite{devries2017improved} to train the ResNet-18 model~\cite{he2016deep}. Weight decay is $5\times 10^{-4}$, learning rate is 0.1, and is decayed by a factor of 5 after 120, 240, and 320 epochs and the model is trained for 400 epochs. For all experiments, we use standard random cropping and horizontal flipping augmentations, and train with Nesterov~\cite{nesterov1983method} accelerated SGD with 0.9 momentum and batch-size of 128.

\subsection{PyramidNet-200 on CIFAR100}
We use the same training setup as in~\cite{yun2019cutmix}, namely, PyramidNet~\cite{han2017deep} initialized with depth 200 and $\tilde{\alpha}=240$, weight decay of $10^{-4}$, learning rate of 0.25 that is decayed by a factor of 10 after 150 and 225 epochs. For all experiments we use standard random cropping and horizontal flipping augmentations and train with Nesterov accelerated SGD for 300 epochs with 0.9 momentum and batch-size of 64.

\subsection{BinaryNet on CIFAR100}
We implemented Binary-Weight~\cite{courbariaux2015binaryconnect,rastegari2016xnor} ResNet-18 architecture, where all weights (with the exception of the first and the last layers) are represented with 1-bit. We use the binary architecture in~\cite{rastegari2016xnor}, and training setup in~\cite{martinez2020training}, namely, weight decay is zero, learning rate is $2 \times 10^{-4}$ that is decayed by a factor of 10 after 150 and 250 epochs. For all experiments, we use standard random cropping and horizontal flipping augmentations and train with ADAM optimizer~\cite{kingma2014adam} for 350 epochs with 0.9 momentum and batch-size of 128. The implementation is the same as~\cite{pouransari2020least}.

\subsection{ResNet on ImageNet}
We use the training setup introduced in~\cite{he2019bag}: the weight decay is $10^{-4}$ and learning rate is linearly warmed-up during the first 5 epochs from 0.1 to 0.4, and then decayed to 0 by a cosine function. For all experiments, we use SGD with Nesterov with batch-size of 1024, and apply standard data augmentations during training: random crop and resize to 224$\times$224, random horizontal flipping, color jittering, and lightening. We resize the images to 256$\times$256 followed by a center cropping to 224$\times$224 during test. We use 300 epochs to train all models similar to~\cite{yun2019cutmix}. We use regular ResNet-50 and ResNet-101 architectures~\cite{he2016deep} (not the D variant introduced in~\cite{he2019bag}).
To reduce under-fitting for XCL, we used $10\times$ smaller weight decay ($10^{-5}$). Note that using a reduced weight decay did not help for other methods.

\subsection{XCL with GAN-Generated Synthetic Data}
\oncelchange{Transfer-set using GAN can be generated online or offline. In the online setting, in every batch, we generate a new set of images via GAN. This setting is used when GPU memory is sufficient to generate a batch of samples in parallel with distillation (e.g. for CIFAR100 benchmark). In the offline setting, we sample a large  dataset using GAN, use the teacher to obtain soft labels, and store the extended transfer-set to be used in distillation.}

\oncelchange{For the XCL-GAN in 2D gaze estimation experiment, we used the conditional GAN in~\cite{shrivastava2017learning} to sample eye images with random orientations in offline setting, adding $\sim485$k examples to the original transfer-set.}

\hadi{For the classification tasks, we used BigGAN~\cite{brock2018large} to generate samples when using XCL-GAN. BigGAN architecture is a conditional GAN, where an embedding vector $\bm{v}_i$ is trained for each class $i$. At inference time, the generative model $G$ gets a latent variable $\bm{z}$ and an embedding vector $v_i$ to generate a random sample $G(\bm{z}; \bm{v}_i)$ from class $i$. To sample with mixed class vectors (between two classes $i$ and $j$) as in Section 4.2, we interpolate the embedding vectors: $\bm{v} = \lambda  \bm{v}_i + (1-\lambda)\bm{v}_j$. We then generate a mixed sample $G(\bm{z};\bm{v})$.}

\oncelchange{In CIFAR100 experiments, we used the online setting. In ImageNet experiments, we used the offline setting and sampled a transfer-set of 1M images using mixed-class labels. At each iteration of the training we sample half of the samples from the generated transfer-set and the other half from the original data points.}



\section{Results of ResNet-50 Training on ImageNet}
The results are shown in Table \ref{tab:results_imagenet_res50}. Compared to the \oncelchange{standard} KD we observe that XCL obtains {\bf 33$\%$} reduction in the teacher-student accuracy gap.

\begin{table}[hbt]
\begin{center}
\resizebox{\columnwidth}{!}{
\begin{tabular}{lcccccc}
  \hline
method&
\begin{tabular}{@{}c@{}}val\\ top-1\end{tabular}&
\begin{tabular}{@{}c@{}}val\\ top-1 gap\end{tabular}&
\begin{tabular}{@{}c@{}}val\\ top-5\end{tabular}&
\begin{tabular}{@{}c@{}}V2-A\\ top-1\end{tabular}&
\begin{tabular}{@{}c@{}}V2-B\\ top-1\end{tabular}&
\begin{tabular}{@{}c@{}}V2-C\\ top-1\end{tabular}\\
  \hline
  \hline
ERM&
77.3&
\multirow{3}{*}{N/A}&
93.6&
74.5&
65.6&
79.4\\
+MixUp~\cite{zhang2017mixup}&
77.8 &
&
93.9 &
74.9 &
66.4 &
79.7 \\
+CutMix~\cite{yun2019cutmix}&
78.7 &
&
94.3 &
75.5 &
66.9 &
80.2 \\
  \hline
KD~\cite{hinton2015distilling}&
79.2 &
2.4 (-)&
94.3 &
75.6 &
67.3 &
80.6 \\
XCL-Mix&
{\bf 80.0 }&
{\bf 1.6 \rbetter{33} }&
{\bf 95.0 }&
{\bf 77.2 }&
{\bf 68.2 }&
{\bf 81.3 }\\
  \hline
  \end{tabular}
  \caption{Accuracies ($\%$) of the ResNet-50 model trained on the ImageNet dataset. Teacher is a ResNet-152-D model trained with CutMix (top-1 acc. = 81.6$\%$). The std of XCL val top-1 is $\simeq 0.1$.}
  \label{tab:results_imagenet_res50}}
  \end{center}
\end{table}

\section{Effect of Teacher \hadi{Model}}
In this section, we explore alternative choices of the teacher $\tau$.
We use XCL-Mix with the same training setup as Section~5.2.
We compare alternative choices of teacher in Table \ref{tab:teacher}. Each teacher is an ensemble of 8 instances of the given model, trained with different initializations. We observe a general trend that a more accurate teacher results in a more accurate student. \cite{muller2019does} observed that when teacher is trained with Label Smoothing (LS), it is more accurate, but can transfer less knowledge to the student. We observe that using XCL, a teacher trained with LS not only is more accurate but also trains a more accurate student.

\begin{table}[tb]
\centering
\resizebox{0.8\columnwidth}{!}{
\begin{tabular}{llll}
  \hline
teacher&
\begin{tabular}{@{}l@{}}$\hat{H}$\\($\%$)\end{tabular}&
\begin{tabular}{@{}l@{}}teacher\\top-1 ($\%$)\end{tabular}&
\begin{tabular}{@{}l@{}}student\\top-1 ($\%$)\end{tabular}\\
  \hline
  \hline

ResNet-18&
23.4&
81.4&
80.2 $\pm$ 0.1\\

+LS $\varepsilon$=0.1&
44.9&
82.5&
80.9 $\pm$ 0.2\\

+MixUp&
31.0&
83.1&
81.1 $\pm$ 0.2\\

+CutMix&
28.2&
84.6&
83.1 $\pm$ 0.2\\

Pyr.+CutMix&
15.1&
87.5&
83.8 $\pm$ 0.1\\

  \hline
  \end{tabular}}
\caption{Analysis of different teachers. Each teacher is an ensemble of 8 models shown in each row.}
  \label{tab:teacher}
\end{table}

\section{Analysis of Student Size}
\begin{figure}[bt]
\centering
    \subfloat[\label{fig:student_size_full}]{
        \centering
        \includegraphics[width=0.9\textwidth]{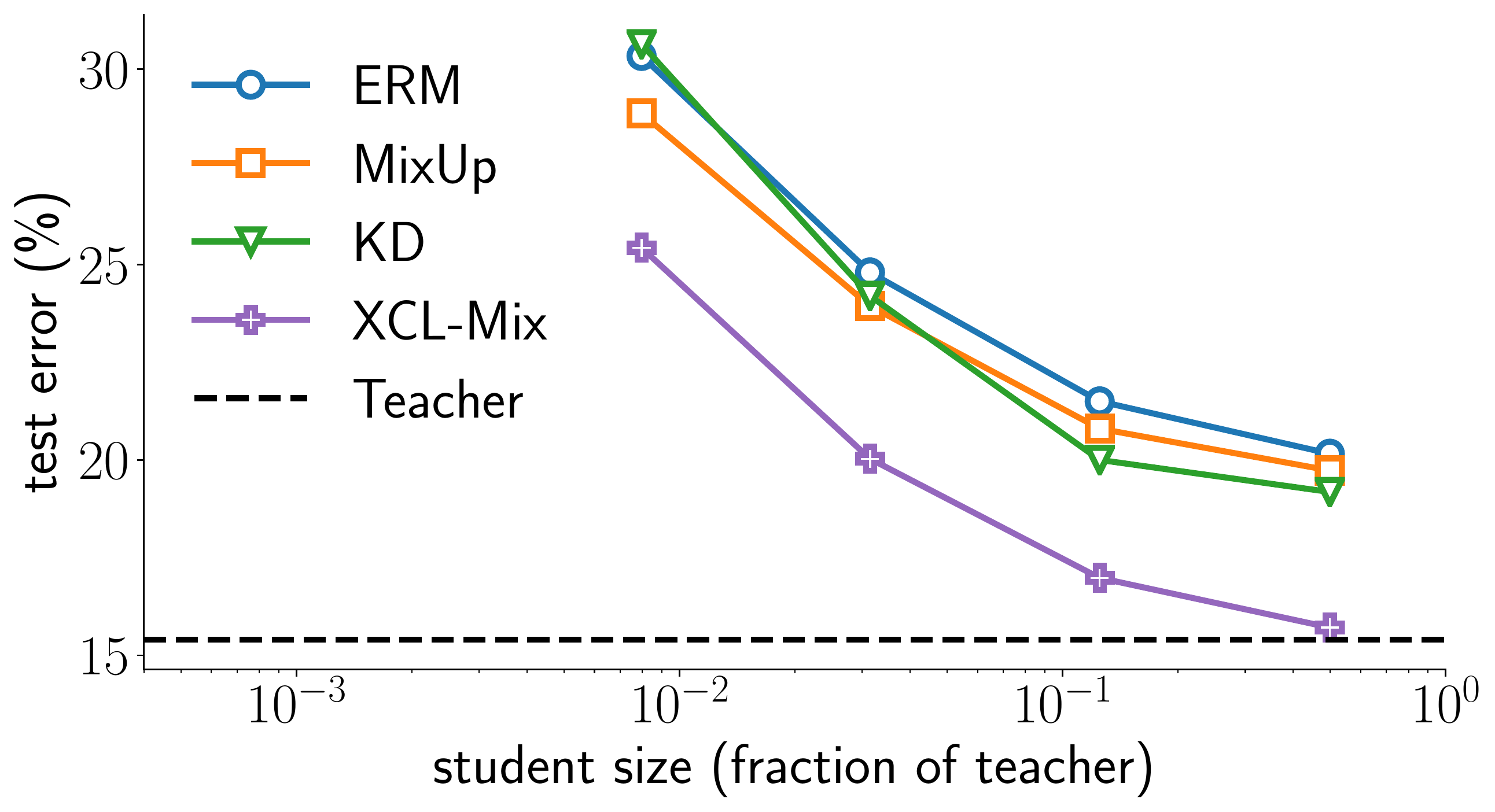}
        }\\
    \subfloat[\label{fig:student_size_binary}]{
        \centering
        \includegraphics[width=0.9\textwidth]{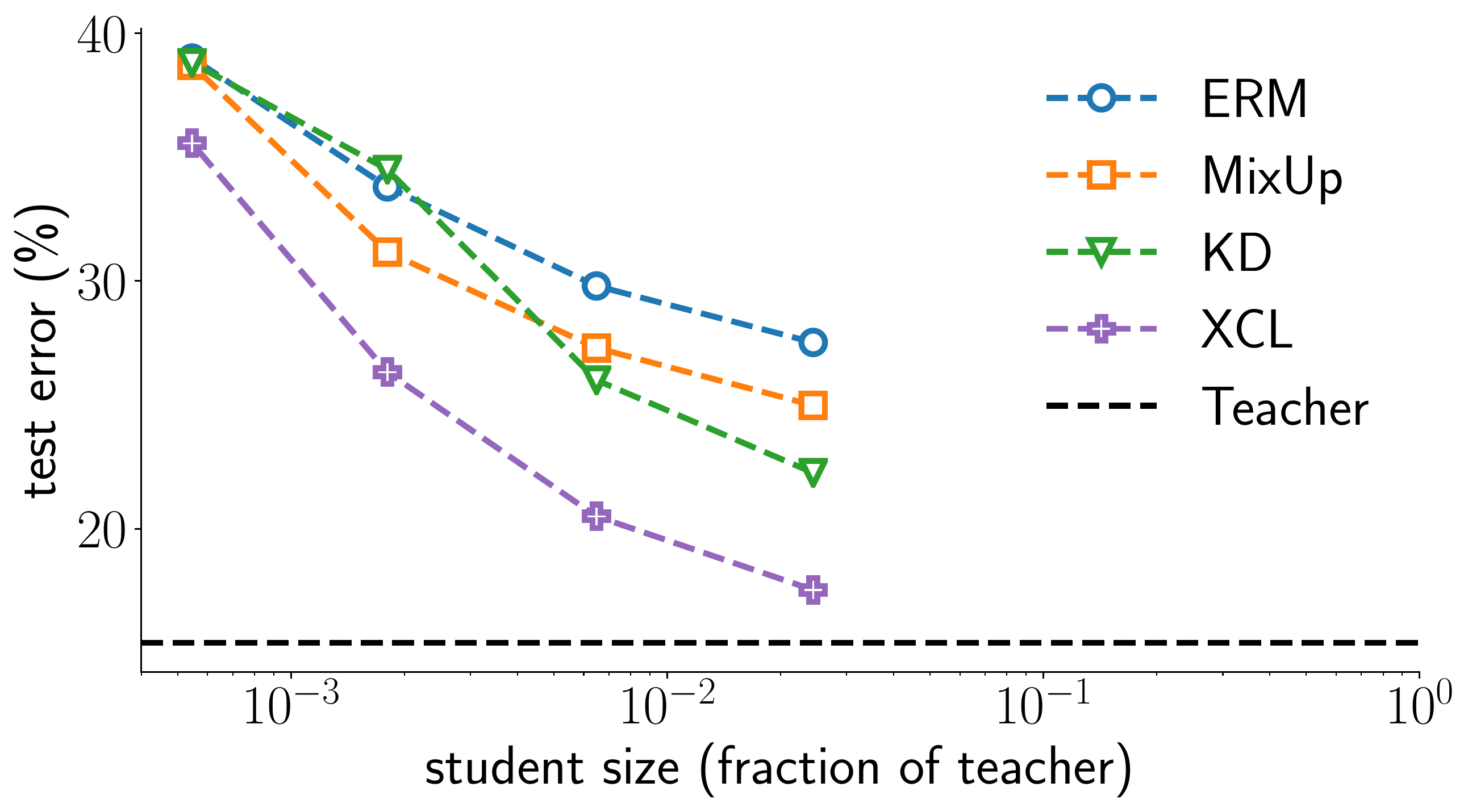}
        }
\caption{\new{Test error as a function of student model size for (a) full-precision and (b) binary training.}}
\label{fig:student_size}
\end{figure}

We investigate the effect of student size on the performance of XCL and other baseline methods. Figures \ref{fig:student_size_full} and \ref{fig:student_size_binary} show the student model accuracy as a function of model size (changed by scaling the channel widths) for both a full precision student and a binary quantized student\hadi{, respectively}. \oncelchange{As seen, XCL consistently outperforms ERM and MixUp augmentation, as well as the standard KD which uses the empirical distribution as the transfer-set.}
\hadi{It is also worth mentioning that the binary model has a better error-size trade-off curve compared to the full precision model.}

\section{Label Smoothing}
Label smoothing (LS)~\cite{szegedy2016rethinking} with parameter $\varepsilon$ replaces ground truth labels with:
\begin{equation}\label{eqn:label_smoothing}
y^{j} = \quad1- \varepsilon \quad \mbox{ if: } \ j \mbox{ is the correct class} \quad \mbox{else:} \ \frac{\varepsilon}{c-1} 
\end{equation}
This method artificially increases label entropy.
The results of ERM training with LS are reported in Table \ref{tab:ls}. \hadi{For all experiments we use the CIFAR100 dataset and ResNet18 architecture as described in Section 5.2.}
Using $\epsilon=0.1$, LS achieves $1.5\%$ improvement over the baseline. Note that, finding an optimal $\epsilon$ requires extensive hyper-parameter tuning. XCL naturally obtains smooth labels, and without hyper-parameter tuning obtains significant accuracy improvement (by 3.8$\%$) compared to the best LS.

\begin{table}[bt]
\centering
\resizebox{0.6\columnwidth}{!}{
\begin{tabular}{lll}
  \hline
$\varepsilon$&
$\hat{H}$ ($\%$)&
top-1 ($\%$)\\
  \hline
  \hline

0.1&
17.0&
79.3 $\pm$ 0.3\\

0.18&
28.2&
78.8 $\pm$ 0.2\\

0.4&
54.5&
78.0 $\pm$ 0.3\\  

0.8&
90.7&
78.3 $\pm$ 0.2\\

  \hline
  \end{tabular}
    \caption{Effect of label smoothing on ERM.}
  \label{tab:ls}}
\end{table}

\section{Knowledge Distillation with Temperature Scaling} 
In KD~\cite{hinton2015distilling}, logits of the student and the teacher are inversely scaled by a temperature parameter $T$ before softmax probabilities are computed. This smoothing strategy can slightly improve the knowledge distillation accuracy (+1.2$\%$ compared to KD without temperature scaling). \hadi{We use the CIFAR100 dataset and ResNet18 architecture as described in Section 5.2.} Results are reported in Table \ref{tab:temp_kd}.

We observe that XCL is not sensitive to temperature (Table \ref{tab:temp_xcl}). Note that finding an optimal $T$ requires extensive hyper-parameter tuning. XCL does not require hyper-parameter tuning, and compared to the best KD with temperature scaling reduces the accuracy gap by 59$\%$.

\begin{table}[bt]
\centering
\resizebox{0.6\columnwidth}{!}{
\begin{tabular}{cll}
  \hline
T&
$\hat{H}$ ($\%$)&
top-1 ($\%$)\\
  \hline
  \hline

1&
10.5&
80.0 $\pm$ 0.2\\

1.5&
52.8&
79.9 $\pm$ 0.3\\

2&
81.2&
80.2 $\pm$ 0.1\\

5&
99.2&
81.2 $\pm$ 0.3\\

10&
99.9&
81.1 $\pm$ 0.3\\

  \hline
  \end{tabular}
    \caption{Effect of temperature on KD.}
  \label{tab:temp_kd}}
\end{table}

\begin{table}[bt]
\centering
\resizebox{0.6\columnwidth}{!}{
\begin{tabular}{cll}
  \hline
T&
$\hat{H}$ ($\%$)&
top-1 ($\%$)\\
  \hline
  \hline

1&
28.2&
83.1 $\pm$ 0.2\\

1.5&
65.0&
83.0 $\pm$ 0.1\\

2&
85.8&
83.1 $\pm$ 0.2\\

5&
99.2&
83.2 $\pm$ 0.2\\

10&
99.9&
83.1 $\pm$ 0.1\\

  \hline
  \end{tabular}
\caption{Effect of temperature on XCL.}
  \label{tab:temp_xcl}}
\end{table}

\end{document}